\documentclass{article} 
\usepackage{times}  
\usepackage{helvet}  
\usepackage{courier}  
\usepackage[hyphens]{url}  
\usepackage{graphicx} 
\usepackage{ijcai16}

\usepackage{hyperref}
\usepackage{caption} 
\DeclareCaptionStyle{ruled}{labelfont=normalfont,labelsep=colon,strut=off} 
\usepackage{amsmath}
\usepackage{amssymb}
\usepackage{booktabs}

\usepackage{standalone}
\usepackage{subcaption}
\usepackage{color, colortbl}
\definecolor{Gray}{gray}{0.8}
\usepackage{placeins}

\DeclareMathOperator*{\argmax}{argmax}

\usepackage{algorithm, algorithmicx}
\usepackage[noend]{algpseudocode}

\DeclareRobustCommand{\okina}{%
  \raisebox{\dimexpr\fontcharht\font`A-\height}{%
    \scalebox{0.8}{`}%
  }%
}

\title{Reliable Multi-Object Tracking in the Presence of Unreliable Detections}
\author{
	Travis Mandel$^1$,
	Mark Jimenez$^1$,
	Emily Risley$^1$, 
	Taishi Nammoto$^1$, \\
	\textbf{Rebekka Williams}$^2$,
	\textbf{Max Panoff}$^3$,
	\textbf{Meynard Ballesteros}$^1$,
	\textbf{Bobbie Suarez}$^1$\\
	$^1$University of Hawa{\okina}i at Hilo\\
	$^2$Mount St. Mary's University\\
	$^3$University of Florida\\
	\{tmandel, markjime, risleye, nammoto, meynardb, suarezb\}@hawaii.edu,\\
	r.i.williams@email.msmary.edu,
   m.panoff@ufl.edu
}
\begin{document}



\maketitle







\begin{abstract}
Recent multi-object tracking (MOT) systems have leveraged highly accurate object detectors; however, training such detectors requires large amounts of labeled data. Although such data is widely available for humans and vehicles, it is significantly more scarce for other animal species.  We present Robust Confidence Tracking (RCT), an algorithm designed to maintain robust performance even when detection quality is poor.  In contrast to prior methods which discard detection confidence information, RCT takes a fundamentally different approach,  relying on the exact detection confidence values to initialize tracks, extend tracks, and filter tracks. In particular, RCT is able to minimize identity switches by efficiently using low-confidence detections (along with a single object tracker) to keep continuous track of objects.  To evaluate trackers in the presence of unreliable detections, we present a challenging real-world underwater fish tracking dataset, FISHTRAC. In an evaluation on FISHTRAC as well as the UA-DETRAC dataset, we find that RCT outperforms other algorithms when provided with imperfect detections, including state-of-the-art deep single and multi-object trackers as well as more classic approaches.   Specifically, RCT has the best average HOTA across methods that successfully return results for all sequences, and has significantly less identity switches than other methods. 
\end{abstract}


\section{Introduction}
Multi-object tracking (MOT) is a longstanding computer vision problem in which the goal is to keep track of the identities and locations of multiple objects throughout a video.  A  popular MOT approach is tracking-by-detection~\cite{luo2020multiple}, in which an object detector is first run on every frame, and those detections are fed as input to a MOT algorithm.  Convolutional neural networks (CNNs) have led to the creation of highly accurate detectors~\cite{ciaparrone2020deep}, thus spurring the development of approaches that rely heavily on these high-quality detections, e.g.~\cite{bochinski2017high}.  

Training such highly accurate detectors requires significant labeled data.  The majority of the MOT literature has focused on tracking pedestrians and vehicles~\cite{luo2020multiple,wen2020ua}, two settings in which labeled data is plentiful. 
However, in specialized tracking scenarios we may have considerably less data; for instance, tracking a new species of insect, or tracking fish off the coast of a tropical island.
With limited training data, even the best detectors will have limited performance.  An ideal tracking algorithm would be able to perform robustly even given an imperfect detector~\cite{solera2015towards}, but it is still not clear how to accomplish this.

One alternative which has gained popularity recently is to forgo tracking-by-detection altogether and train an end-to-end approach that simultaneously learns to detect and track objects of interest~\cite{feichtenhofer2017detect,sun2020simultaneous,zhou2020tracking}.  Although useful in many situations, this approach requires a large dataset of videos labeled with track information. In the settings we study, there is little to no labeled video data needed for end-to-end tracking approaches. Indeed, even properly labeled still image data needed to train a standard object detector may be fairly scarce, greatly increasing the difficulty of the problem compared to the standard MOT setting.



We have found that even when (pretrained) CNN detectors are trained on little data, they often are still able to predict  the general location of objects in the scene, albeit sometimes with very low confidence and many false positives.  However, the traditional MOT pipeline discards most of this information, first filtering out the low-confidence detections, and thereafter discarding the detection confidence values~\cite{wen2020ua}.  Ideally, a tracker could make use of the full \textit{unfiltered} set of detections to achieve more robust performance.  Unfortunately, removing this filtering step greatly increases the computational burden, and requires algorithms to cope with extremely noisy input.  Due to these challenges, we are not aware of any tracking-by-detection approaches capable of efficiently handling an unfiltered set of detections.

Therefore, we present Robust Confidence Tracking (RCT), an algorithm which tracks efficiently and robustly given unfiltered detections as input. The key idea behind RCT is that, instead of discarding detection confidence values, we can use these values to guide the tracking process, using lower-confidence detections only to ``fill in gaps'' between higher-confidence detections.  Specifically, RCT uses detection confidence in three ways: To determine where to best initialize tracks,  probabilistically combined with a Kalman Filter to optimally extend tracks, and to filter out low-quality tracks. Alongside this, RCT incorporates the Median Flow single object tracker (SOT) and some simple heuristics for track trimming and joining to achieve excellent performance, even compared to more complicated and resource-intensive deep tracking methods.
To test trackers such as RCT in challenging scenarios where  data is scarce, datasets of common objects do not suffice.  
Therefore,  we present a new,  challenging real-world fish tracking dataset, FISHTRAC.  We conduct a comprehensive evaluation 
on both FISHTRAC as well as the UA-DETRAC~\cite{wen2020ua}  vehicle dataset (using a low-accuracy detector). 

\section{Problem Setup}
We consider \textbf{offline} multi-object tracking problems within a tracking-by-detection framework, where the goal is to track all objects of a desired class $\ell$ throughout a video sequence. Note that $\ell$ is often known to be of practical importance, but in the settings we consider may be rare enough that accurate detection is difficult.
Specifically, we assume there exists a video $\mathcal{V}$ with $N$ frames $v_1,\dots,v_N$ and a detector $\mathcal{D}$ which outputs detections on each frame $d_1,\dots,d_N$.  Each $d_i$ is a set containing tuples $b = (x,y,w,h,c)$ denoting the detected bounding box and its confidence $0 \leq c \leq 1$ that the box corresponds to an instance of an object of class $\ell$ (here we use $b \in \ell$ to denote the case that a box is a member of the class $\ell$).  The goal of the tracking algorithm $\mathbb{T}$ is to produce an optimal set of tracks $T = {T_1,\dots,T_K}$ where each track $T_j$ consists of a list of tuples $t_{j,f} = (x,y,w,h,c)$ where $f \in [1,N]$ is the frame number.  

\section{Related Work}

\textbf{Multi-object Tracking Datasets and Codebases:}
%
There are several public MOT datasets, see Table~\ref{tab:datasetcompare}.  However, tracking fish in natural underwater scenarios is a challenging and understudied problem which is not well-represented by existing datasets.\footnote{Note that our assessment of 2 real-world fish videos for TAO~\cite{dave2020tao} is based only on examining the TAO train and validation dataset as the test dataset is not yet fully released. Similarly, our assessment of 2 real-world fish videos for OVIS~\cite{qi2021occluded} is based only on examining the OVIS train dataset as the validate and test datasets are not yet fully released.} Although some datasets do include fish data, the video is usually of artificial settings such as aquarium tanks, which greatly simplifies the tracking problem.   The one significant exception is Fish4Knowledge/SeaCLEF~\cite{jager2016seaclef,jager2017visual,kavasidis2012semi,kavasidis2014innovative}, however that dataset suffers from several problems, including low image quality 
and low FPS (5 FPS).
Indeed, most of the datasets with more variety have sacrificed FPS (e.g. 1 FPS for TAO~\cite{dave2020tao}), and some such as TAO also have incomplete annotations, making comphrensive evaluation difficult. Low FPS is a particularly poor choice for fish tracking, since fish move and change direction rapidly. 
Our FISHTRAC dataset contains high-quality (at least 1920x1080) video of real-world underwater fish behavior, and is completely annotated at 24 FPS.  While not as diverse or large as datasets like TAO~\cite{dave2020tao}, FISHTRAC fills an important gap by helping shed light on a highly challenging real-world application.~\footnote{The public release of the dataset is forthcoming.}

\begin{table*}
  \begin{center}
	  \caption{A comparison of public MOT datasets. FPS refers to the annotation FPS.}

    \label{tab:datasetcompare}
		\resizebox{\textwidth}{!}{
    \begin{tabular}{p{3.2cm}|p{1cm}|p{2cm}|p{1cm}|p{1.5cm}|p{3cm}|p{1.3cm}|p{2cm}} %
      \textbf{Dataset} & \textbf{Num Videos} & \textbf{\# ``In the wild'' fish videos}  & \textbf{FPS} & \textbf{Min resolution} & \textbf{Provides unfiltered detections?} & \textbf{Complete labels?} & \textbf{\# MOT algs in codebase} \\\specialrule{2.5pt}{1pt}{1pt}
			UA-DETRAC~\cite{wen2020ua} & 100  & 0  & 24 & 960x540 & No & No & 8 \\\hline
			KITTI~\cite{geiger2013vision} & 40 & 0 & 10  & 1242x375 & No & No & 0 \\\hline
			TAO~\cite{dave2020tao} & 2,907 & 2  &  1  & 640x480 & N/A & No & 1 \\\hline
			MOT20~\cite{dendorfer2020mot} & 8 & 0   & 30 & 1173x880 & No & Yes & 0 \\ \hline
			YTVIS 2021~\cite{yang2019video} & 3,859 & 2   &  5 & 320x180 & N/A & Yes & 0 \\\hline
			OVIS~\cite{qi2021occluded} & 901 & 2   &   3-6  & 864 x 472 & N/A & Yes & 0 \\\hline
			SeaCLEF   \cite{jager2017visual,kavasidis2012semi,kavasidis2014innovative} & 10 & 10   & 5 & 320x240 & N/A & No & 2 \\
			\specialrule{2.5pt}{1pt}{1pt}
			FISHTRAC (Ours) & 14 & 14 & 24 & 1920x1080 & Yes & Yes & 16 \\ \hline

    \end{tabular}}
  \end{center}
\end{table*}

Additionally, there is currently a lack of MOT codebases that facilitate comparison to other methods.   Leaderboards such as  MOTChallenge~\cite{milan2016mot,dendorfer2020mot} are the predemoninant way to compare trackers, but this does not allow one to compare algorithms on new videos or detections. 
Running trackers on a new dataset takes substantial implementation time and effort (converting formats, handling very slow trackers, etc.). 
The UA-DETRAC~\cite{wen2020ua} codebase allows one to compare 8 trackers, but it is intended for only a single dataset and is based on proprietary (paid) software (MATLAB).   We present a heavily modified version of the DETRAC code which is adapted to open-source technologies and contains everything needed to run 16 trackers on fish data, car data, or a new dataset.~\footnote{The public release of the codebase is forthcoming.}  

\textbf{Fish tracking:}  
Work in real-world fish tracking has been relatively scarce due to lack of suitable datasets. Most attention has focused on artificial settings, such as tracking Zebra Fish in a glass enclosure~\cite{pedersen20203d,romero2019idtracker}.
One exception is J{\"a}ger et al.~\shortcite{jager2017visual}, who developed a custom approach to track fish in real-world scenarios.  
We compare to this tracker in our experiments.

\textbf{Detection Confidence:}
Virtually all tracking-by-detection methods filter detections based on a confidence threshold $h$ and thereafter discard confidence information, e.g. let $d'_f = {b \textrm{ s.t. } c_b \geq h, b \in d_f}$, where $c_b$ denotes the confidence of the box $b$. Indeed, this is enforced by codebases such as UA-DETRAC~\cite{wen2020ua}.  The few exceptions add another threshold to differentiate between high and medium confidence~\cite{bochinski2018extending}, or require modifying detection approaches to expose additional information which may not always be accessible~\cite{verma2003face,breitenstein2009robust}.  Bayesian approaches like JPDA~\cite{fortmann1983sonar,rezatofighi2015joint} incorporate a fixed detection probability, but do not utilize the individual detection confidences.  We are not aware of any MOT algorithms that make use of the detection confidence values associated with each produced bounding box in a manner more sophisticated than just thresholding them and discarding them, a procedure which eliminates the more nuanced information contained in these values.

\section{Robust Confidence Tracking (RCT)}

\begin{figure*}
\center
\includegraphics[width=0.9\textwidth]{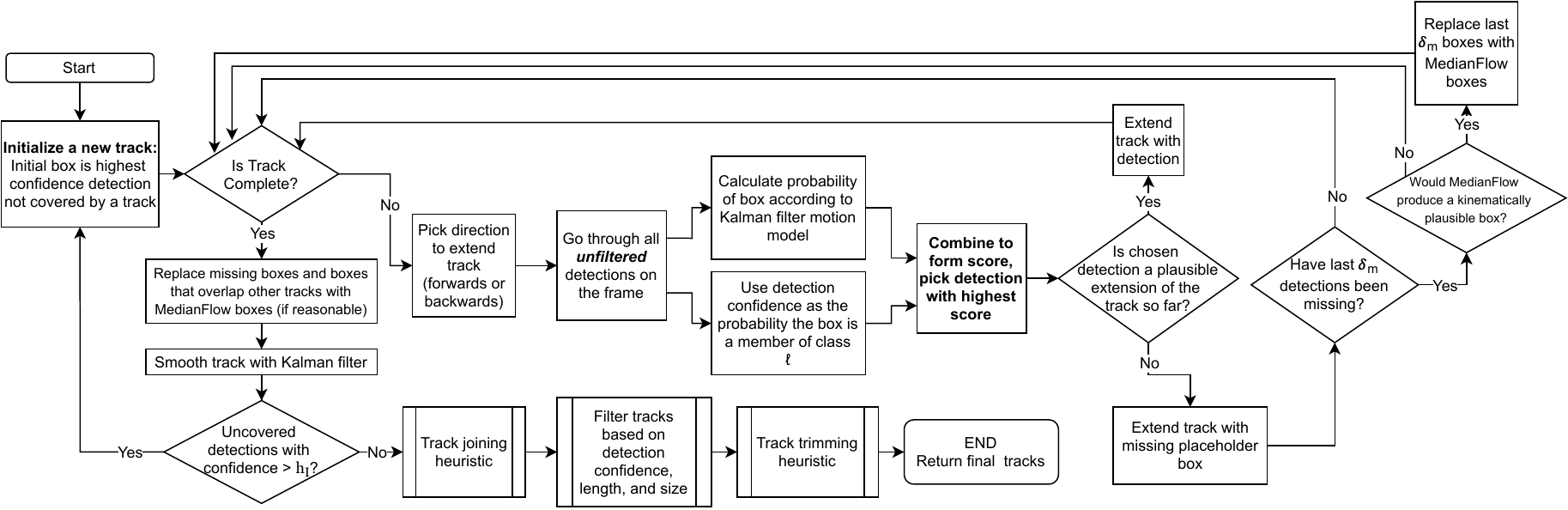}  
\caption{Process diagram of our Robust Confidence Tracking (RCT) algorithm. }
\label{fig:rct_alg}
\end{figure*}

Our Robust Confidence Tracking (RCT) algorithm contains four components: Initializing tracks based on detection confidence, probabilistically combining detection confidence with motion probability, incorporating a single-object tracker  as a fallback when detections alone are insufficient, and track postprocessing.  Figure~\ref{fig:rct_alg} gives an overview.

\subsection{Initializing tracks based on detection confidence}

Unlike other MOT algorithms, our RCT algorithm uses the detection confidence as a key to distilling the detections into coherent tracks.  For each track $T_j$, RCT chooses the maximum confidence detection (across all frames) for its initial box $I_j$
($I_j = \argmax_{b \in d_1\cup \dots \cup d_N} c_b $)
; we refer to the frame associated with $I_j$ as $f_{I,j}$. To ensure that this detection does not overlap with a previously-used track, RCT excludes boxes from the max where there exists some $w$ such that $t_{w,f_{I,j}} \in d_{f_{I,j}}$ and $|B(t_{w,f_{I,j}}) \cap B(I_j)| > 0$, where the function 
$B$ returns the set of all pixel coordinates that fall within the box. Also, to avoid edge cases, we do not select detections that are near the edge of the screen (we enlarge $I_j$ by $\beta$\% and check that it is still onscreen, where $\beta$ is an RCT parameter). RCT works in a track-wise fashion: once the first track is built (described below), the next highest detection confidence detection is selected as the start frame for the next track, and so on, as long as the initial confidence $c_{I,j}$ is above an RCT threshold parameter $h_I$.  Note that RCT uses detections with confidence $<h_I$ elsewhere.

\subsection{Combining detection confidence and motion}
 Next, RCT initializes a Kalman Filter $k$ with this detection.  The Kalman filter state $s^k$ is a tuple $(x^k,y^k,v^k_x,v^k_y,w^k,h^k)$ where  $v^k_x$ and $v^k_y$ are unobserved (latent) velocities which together form a vector $\vec{v}^k = <v^k_x,v^k_y>$. Let $b^k = (x^k,y^k,w^k,h^k)$ be the box derived from the state. 

From the initial box $I_j$, RCT could extend the track either forward or backward in time, but it does not know which will best help estimate velocity.  To handle this, RCT initially tries both options, and selects the option with best score (as defined below). For clarity we will describe the forward case, the backward case is analogous. 

Given a frame $f$, partial track $j$ and a Kalman filter state $s^k_{f,j}$, RCT must score each detection in $d_f$ to find the best candidate to extend the track.  The Kalman filter probability of the box based on the track so far $P(b' = t_{j,f}|t_{j,f-1},\dots,t_{j,1})$ can be used as a motion model score. Similarly, the Detector $D$ assigns a probabilistic score $P(b' \in \ell|v_f) = c_{b'}$ reflecting the probability the detector believes this object is of the desired  class.  Our goal is to find the joint probability $P(b' = t_{j,f},b' \in \ell|v_f,t_{j,f-1},\dots,t_{j,1})$.  If we make the simplifying assumption that the class and the track assignment are conditionally independent given the past sequence of boxes and frame image, we have: 
\begin{small}
\begin{multline}
P(b' = t_{j,f},b' \in \ell|v_f,t_{j,f-1},\dots,t_{j,1}) \\= P(b' = t_{j,f}|v_f,t_{j,f-1},\dots,t_{j,1})P(b' \in \ell|v_f,t_{j,f-1},\dots,t_{j,1})
\end{multline}
\end{small}
Since our detector gives $P(b' \in \ell|v_f,t_{j,f-1},\dots,t_{j,1}) = P(b' \in \ell|v_f) = c_{b'}$, and our Kalman filter assumes $P(b' = t_{j,f}|v_f,t_{j,f-1},\dots,t_{j,1}) = P(b' = t_{j,f}|s^k_{j,f})$, the joint probability can be calculated as: 
\begin{small}
\begin{equation}
P(b' = t_{j,f},b' \in \ell|v_f,t_{j,f-1},\dots,t_{j,1}) = c_{b'} P(b' = t_{j,f}|s^k_{j,f}). \label{eq:rctcore}
\end{equation}
\end{small}
RCT uses equation~\eqref{eq:rctcore} to a score a detected box based on both detection confidence and motion model score. However, this does not tell us when none of the detected boxes on a certain frame are a reasonable extension of the track - a situation that arises frequently with an imperfect detector.  To do so, RCT checks two criteria.  First, RCT checks whether the the center point of the chosen detection $b'$ is contained within the box derived from the Kalman filter state, specifically $C(b') \in B(b^k_{j,f})$ where $C$ is a function  that returns the geometrical center of the box. If not, it is likely not a kinematically plausible extension of the track.\footnote{We optimize by only considering detections that overlap $b^k_{j,f}$.}  Next, RCT checks if $P(b' = t_{j,f}|s^k_{j,f}) \geq  P(b' = t_{j,f-1}|s^k_{j,f-1})$, in other words, whether the previous detection is more likely under the current Kalman filter state than the previous Kalman filter state: if not, it is likely moving in the wrong direction.  If the detection is rejected due to either of the above reasons, RCT sets  $t_{j,f}$ to a placeholder value indicating a missing observation.  Otherwise, RCT sets $t_{j,f} = b'$, and marks $b'$ so that it cannot be re-used in another track.
 
After $\delta$ ($\delta$ is an RCT parameter) iterations extending the track in both directions, RCT then switches to a single direction (forward, then backward) as the estimate of the velocity is likely sufficiently accurate. RCT stops this process when the current box is more offscreen  than the last box, setting the rest of the $t_j$ to missing since the object is offscreen. 

The Kalman filter is used to perform one final smooth at the end, letting $t_{j,f} = b^k_{j,f}$ to smooth out any noise in the track and replace missing observations with inferred boxes.\footnote{Specifically, RCT runs a (forward-backward) smoothing pass separately on frames $f' < f_{I,j} + \delta$  and $f' > f_{I,j} - \delta$, where the $\delta$  extra frames past the start frame are used as context.}

\subsection{Incorporating a single object tracker}
The aforementioned approach forms the core of the RCT algorithm; however, the Kalman filter assumes linear motion if we do not find a matching detection, which performs poorly when motion is complex.  Therefore, RCT uses a SOT algorithm as a fallback option if no reasonable detections can be found.
Specifically, we use the MedianFlow tracker~\cite{kalal2010forward}: MedianFlow has been successfully used in past MOT approaches~\cite{bochinski2018extending}; a strength is that it can determine when it has lost track of an object.  As with the Kalman Filter, we initialize the MedianFlow tracker on frame $I_f$ and update it in both the forwards and the backwards directions.

We observed that the Kalman filter could overcome a short sequence of missing detections or occlusions, while visual information is critical to overcoming a longer sequence of missing detections.  Therefore, RCT switches to MedianFlow only if, when detecting on frame $f$ of track $j$, for all $f' \in \{f,f-1,\dots,f-\delta_m\}$ $t_{j,f'} \not\in d_{f'}$ (i.e. the last $\delta_m$ frames also have no valid detections), where $\delta_m$ is an RCT parameter.   Additionally, we require that the MedianFlow track is plausible according to our Kalman filter, specifically RCT tests that $C(m_{j,f}) \in B(b^k_{j,f})$, where $m_{j,f}$ is the MedianFlow box on frame $f$ of track $j$.  If these conditions are met, and MedianFlow did not report a tracking failure, RCT sets $t_{j,f'} = m_{j,f'}$ for $f' \in \{f,f-1,\dots,f-\delta_m\}$.  In the case where both a MedianFlow box $m_{j,f}$ and an acceptable detected box $b' \in d_j$ are available, and the previous box was MedianFlow ($t_{j,f-1} = m_{j,f-1}$), RCT only sets $t_{j,f} = b'$ if $C(b') \in B(m_{j,f})$ and $C(m_{j,f}) \in B(b')$, which tests whether the detection diverges significantly from the MedianFlow prediction.  If it does, it is likely a spurious detection and RCT keeps using the MedianFlow boxes.

To further reduce the reliance on motion, RCT replaces some boxes with MedianFlow after the track is  built.  If on some track $j$ and frame $f$, either the detection is a missing placeholder or it overlaps with another track  (i.e. there exists some $w \neq j$ such that $t_{w,f} \in d_f$ and $|B(t_{j,f}) \cap B(t_{w,f})| > 0$), we try to see if we can replace $t_{j,f}$ with a better box. First, RCT tries a MedianFlow box: if $|B(m_{j,f}) \cap B(t_{j,f-1})| > 0$, it is a reasonable extension of the track, so we let $t_{j,f} = m_{j,f}$. Else, RCT sets $t_{j,f}$ to indicate a missing observation.

\subsection{Track joining, confidence-based filtering, trimming}
The approach so far can produce tracks that are fragmented - therefore, RCT joins smaller tracks as a postprocessing step. Instead of computationally expensive matching approaches~\cite{dehghan2015gmmcp}, we use a fast and simple greedy heuristic that joins two tracks if they are similar enough in terms of time and motion.  Specifically, RCT examines the time period in which the tracks switched to purely motion. Without loss of generality, let $f_j$ be the last non-motion-box frame of track $j$, and $f_w$ be the first non-motion-box frame of track $w$ (we try every possible ordered pairing of tracks). If $f_j \leq f_w$, then RCT computes the temporal distance as $D_{time} = f_w-f_j$
If $f_j > f_w$, then we require there to be at most two frames $f$ where  $IoU(t_{f,w}, t_{f,j}) < h_u$, where IoU is the intersection over union function and $h_u$ is an RCT parameter. In other words, the track needs to overlap on almost every frame in which there are detections, if so RCT sets $D_{time}=0$.  RCT only considers joining pairs of tracks where $D_{time} < D_{max}$, where $D_{max}$ is an RCT parameter. Next we consider whether the distance reached in that number of frames would be reasonable according to the Kalman Filter.
Specifically, let 
\begin{equation}
v_{max} = \max_{i \in \{w,j\}, f \in \{1,\dots,N\}} \sqrt{\left(v^k_{i,f,x}\right)^2 + \left(v^k_{i,f,y}\right)^2}, 
\end{equation}
which gives the fastest speed it is reasonable for these objects to have under the Kalman filter.  Then we test whether
\begin{equation}
d_{euclid}(C(b_{f_w,w}), C(b_{f_j,j})) \leq D_{time} v_{max},
\end{equation}
where $d_{euclid}$ gives the Euclidean distance. If not, the distance between tracks is too large to reasonably join them. Additionally, RCT checks that it is moving in the right direction: that is, that a Kalman filter initialized on track $w$ and extended to track $j$ would determine that $P(b_{j,f_j}| s^k_{f_j}) \geq P(b_{j,f_j}| s^k_{f_j-1})$. RCT iterates this process, greedily adding tracks until there are no more pairs that meet our join criteria.  After each join, RCT re-smooths the tracks.

Since the detector is low-accuracy, it may be that long sequences of detections occur on objects outside $\ell$ (e.g. coral instead of fish), necessitating track filtering.  RCT relies on detection confidence to filter tracks: Detections of smaller objects tend to be naturally lower confidence, but if a track is both exceptionally long and contains many large boxes, at least some of the detections should be fairly high confidence if it is truly the target class.  To determine if a track $T_q$ qualifies as large and long, we first define a set $\mathcal{T}_l$ consisting of all the high-quality large long tracks. Specifically,  $T_j \in \mathcal{T}_l$ if two conditions are met: $c_{I,j} > h_q$, where $h_q$ is an RCT parameter, and  $S(T_j) \geq  \frac{ \sum_{T_i \in T} S(T_i)}{|T|}$, where $S(T_i) = \sum_{f \in \{1,\dots,N\}} |B(t_{i,f})|$, that is, the a total size (as calculated by summing the box sizes across all frames) greater than the mean across all tracks.   For each $T_i \in \mathcal{T}_l$, RCT computes $S(T_i)$, producing a set of scalar sizes $\mathcal{S}_l$. RCT then fits a Gaussian distribution to the mean and standard deviation of the elements in $\mathcal{S}_l$, the intuition being that the Gaussian distribution captures what sizes are reasonable for large tracks to have in the dataset. If for the track in question $T_q$, $S(T_q)$ is above the 95\% Gaussian tail, and it is low confidence ($c_{I,q} < h_q$), $T_q$ is removed from the track set.  RCT also removes redundant tracks, i.e. where the average IoU between two tracks is greater than RCT parameter $h_u$.

Finally, RCT trims the ends of tracks (which has a large impact on scores, see our ablation study).  Specifically, RCT stops tracking objects when the width of the box is offscreen by more than $\omega$ percent of the frame width, and the height of the box is offscreen by more than $\omega$ percent of the frame height where $\omega$ is an RCT parameter.  When an object is moving offscreen, RCT applies constant acceleration of $\alpha\vec{v}^k$ to the Kalman-derived velocity vector $\vec{v}^k$ to move the track swiftly offscreen, where $\alpha$ is an RCT parameter.  Additionally, to avoid incorrect extrapolation of the tracks by the Kalman filter, RCT trims all boxes that are based on missing Kalman observations at the tail ends of the tracks as long as there are at least $\delta_n$, where $\delta_n$ is an RCT parameter.  

\section{FISHTRAC Dataset}

\subsection{A high-resolution MOT fish dataset}

Real-world underwater fish tracking is a particularly challenging MOT problem. Fish move unpredictably, change appearance, and are frequently occluded.  When video is collected by divers, additional complicated motion and parallax effects arise; additionally, fish often intentionally try to swim away or hide from the diver.  And yet fish tracking is an important task in marine science, for instance to aid in studies of fish behavior, and also has recreational applications.  

\begin{figure}
\includegraphics[width=\columnwidth]{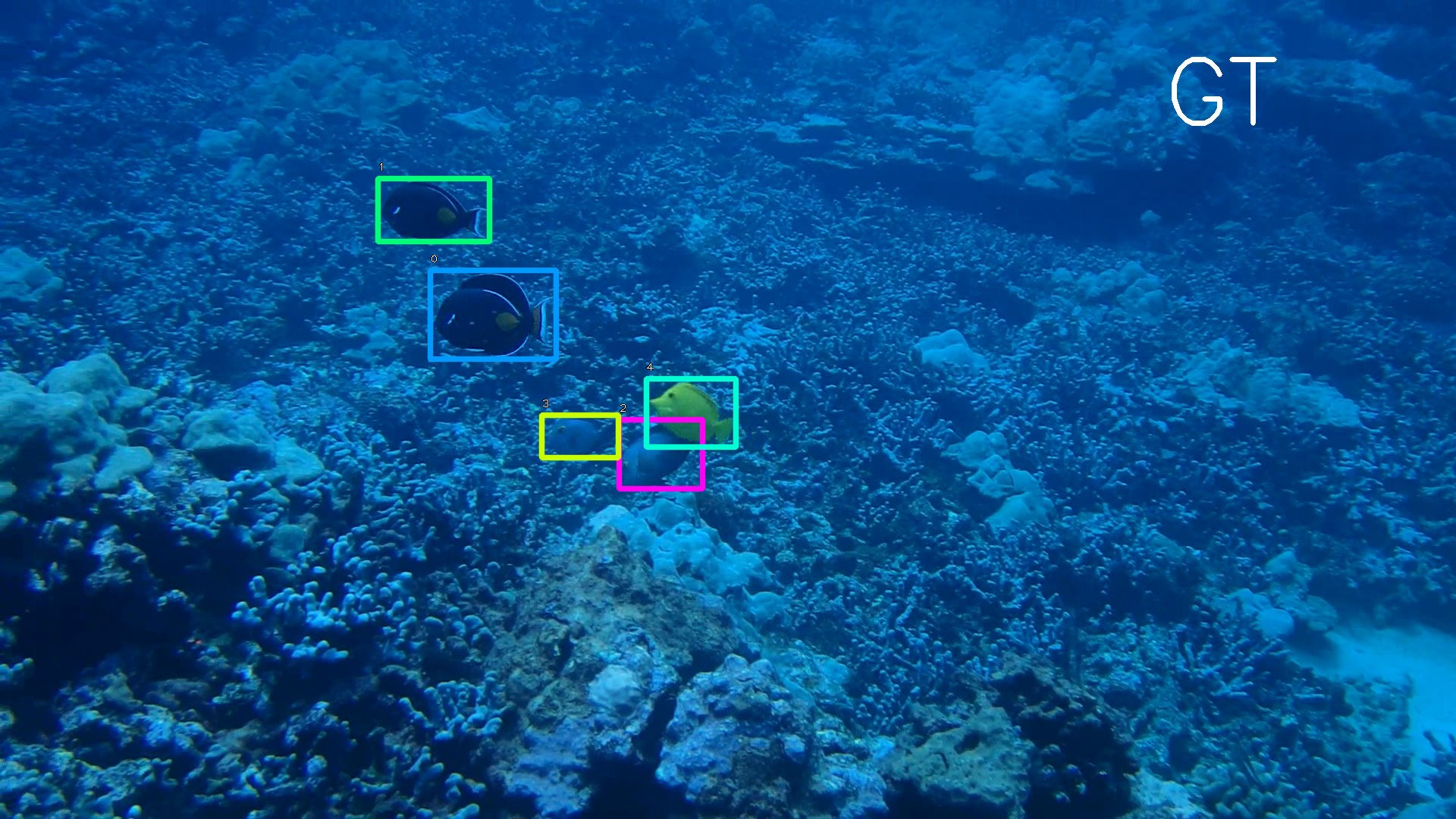}  
\caption{FISHTRAC frame marked with ground truth (GT).}
\label{fig:example_fishtrac}
\end{figure}

We present FISHTRAC, which is, to our knowledge, the first high-resolution fish dataset designed for multi-object tracking.  FISHTRAC contains 14 videos totaling 3,449 fully-annotated frames of real-world underwater video. Annotators were instructed that, if a fish is unambiguously identifiable in at least one frame of video, it should be annotated for all frames that it is believed to be within the camera's Field of View (FOV). 
This results in 131 total individual fish annotated (5-20 per video).   Video is in high-resolution 1920x1080 (or higher) format collected at 24 frames per second, see Figure~\ref{fig:example_fishtrac} for an example.  The videos were collected off the coast of Hawai{\okina}i island, primarily by a SCUBA diver, although we also include a video collected by a snorkeler and a video from a stationary camera. 
To simulate tracking with scarce data, just 3 videos are designated for training, the other 11 are reserved for testing.  Likewise, when training on UA-DETRAC, we use just 3 videos from the train set (MVI\_41073, MVI\_40732, MVI\_40141).

Additionally, we present the FISHTRAC codebase which includes everything needed (conversion/visualization scripts, etc.) to evaluate 16 tracking  algorithms on a new MOT problem.  
Our code is based entirely on free technologies (GNU Octave and Python) and supports Linux.  (The public release of the dataset and codebase are both forthcoming.)

%
\subsection{FISHTRAC object detection}

\begin{figure}
	\centering
	\includegraphics[width=\columnwidth]{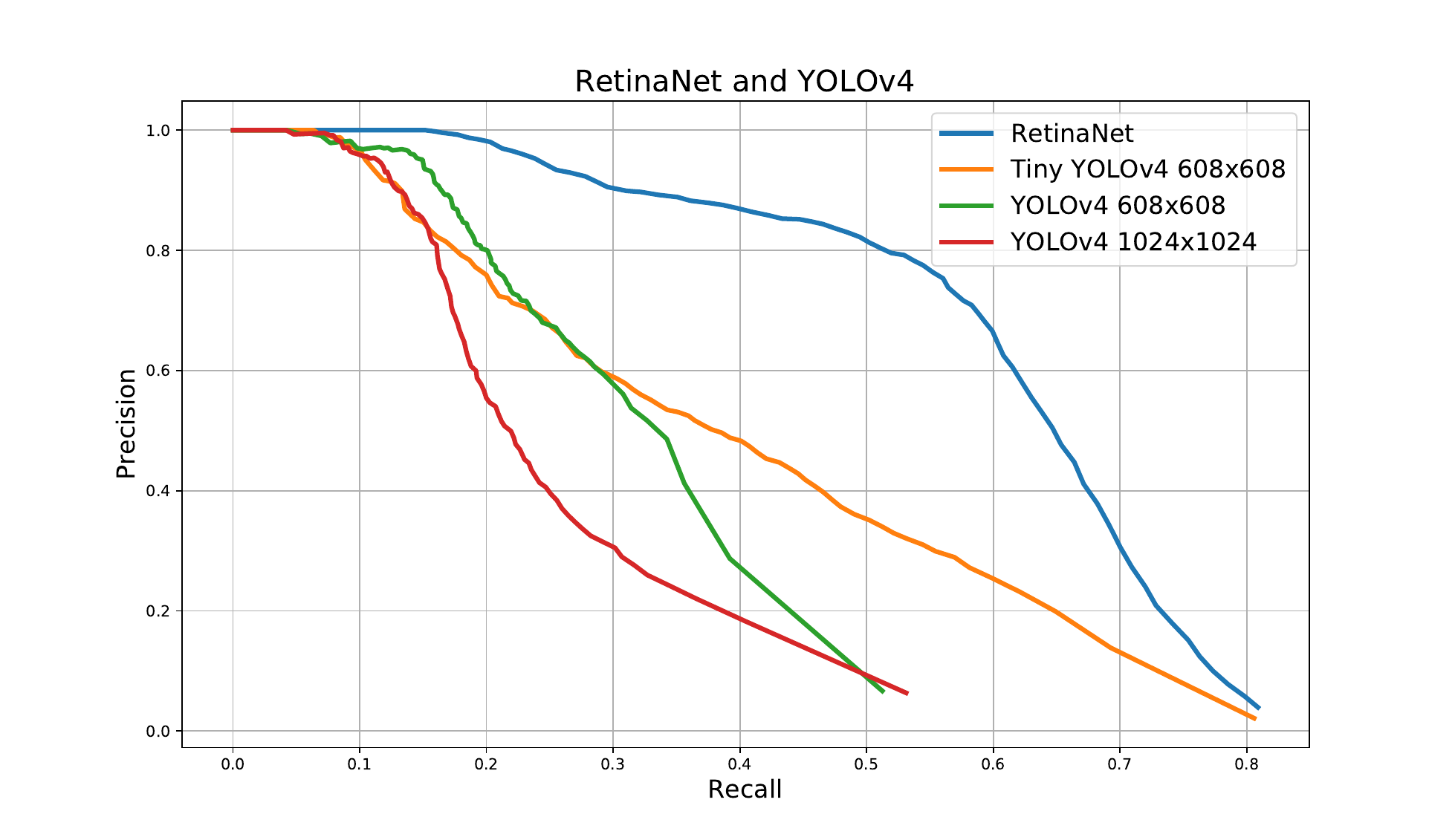}
	\caption{Precision-Recall Curves of RetinaNet compared to YOLOv4 on the FISHTRAC training dataset. }
\label{fig:detectorcomparison}
\end{figure}

\begin{table}
  \begin{center}
    \caption{Precision, recall and mean average precision (mAP) of RetinaNet compared to YOLOv4 on the FISHTRAC train.}
    \label{tab:detectorcomparison}
		\resizebox{\columnwidth}{!}{
    \begin{tabular}{ p{4cm} |l|l|l} %
      \textbf{Algorithm} & \textbf{Precision @ 0.5} & \textbf{Recall @ 0.5} & \textbf{mAP} \\\specialrule{2.5pt}{1pt}{1pt}
		RetinaNet &  \textbf{85.83} & \textbf{42.25} & \textbf{60.41} \\ \hline
		YOLOv4 - 1024x1024  & 76.90 & 16.30 & 25.11 \\ \hline
		YOLOv4 - 608x608  & 83.89 & 18.51 & 30.40 \\ \hline
		YOLOv4 Tiny - 608x608  & 69.94 & 23.64 & 39.32 \\ \hline


    \end{tabular}}
  \end{center}
\end{table}

In order to run a tracking-by-detection MOT pipeline on FISHTRAC, we need to train an object detector; however, this requires significant training data (even after pretraining the network on a general-purpose dataset like ImageNet). Although manually annotating images is one option, that is time-consuming and expensive, and for many applications significant training data is available in large public datasets.   Therefore, to generate training data for our FISHTRAC detectors, we scraped all human-annotated bounding boxes labeled ``fish'' from Google Open Images Dataset~\cite{OpenImages}, one of the largest bounding box datasets. However, this resulted in only 1800 images (many of which were not in real-world underwater environments), which is more limited than the data usually used to train a deep learning model.

Next we examine different object detection approaches - although we wish to study cases where detections are low-quality, it is important to select a detection pipeline that maximizes the quality of detections given our limited training data.
Therefore, we compared state-of-the-art detectors on our FISHTRAC set and selected the one with the best performance. Specifically, we compared the RetinaNet~\cite{lin2017focal} architecture to variants of YOLOv4~\cite{bochkovskiy2020yolov4}. For RetinaNet, we selected a ResNet50 backbone~\cite{he2016deep} pretrained on ImageNet~\cite{deng2009imagenet}, and trained it for 10 epochs. For YOLOv4, we used the officially published model architectures (both full size and Tiny variants).

We followed the official guide  in the GitHub repo\footnote{\url{https://github.com/AlexeyAB/darknet}} to train the YOLO models on our custom objects, using the pretrained MSCOCO weights. The only slight deviation was in how we set the training steps. By the recommended 6000 training steps, loss was still decreasing and was not lower than 1, so per the instructions we increased the number of steps in an attempt to achieve lower loss, specifically trained all YOLO models for approximately 12000 steps. Of these steps we selected the model with the lowest train mAP for evaluation. RetinaNet rescales images to between 800-1000 pixels on each side, whereas YOLOv4 by default rescales its input to 608x608, so we also tried a variant of YOLOv4 with a larger input data size (1024 x 1024). 

We evaluated the various object detection models on the FISHTRAC train dataset, using an IoU threshold of 0.5.  As one can see from the precision-recall curves in Figure~\ref{fig:detectorcomparison} and the scores in Table~\ref{tab:detectorcomparison}, the results show that RetinaNet significantly outperforms the more recent YOLOv4 model on FISHTRAC data, hence justifying our choice of it as the source of our detections.  This is consistent with past work which has shown that RetinaNet performs especially well with very little training data~\cite{bickel2018automated,weinstein2019individual}.  Despite this, the resulting RetinaNet detector still has mediocre performance on FISHTRAC train: at a 0.5 confidence threshold, it has 85.53\% precision and just 42.25\% recall (60.4 mAP) . 

\noindent\textbf{Detection for UA-DETRAC:} For DETRAC, one can train an accurate detector given the ubiquity of vehicle data; but we intentionally trained on limited data to realistically simulate poor quality detections. Specifically, we trained RetinaNet with 200 car images from  Google Open Images.  We then trained the same RetinaNet architecture used in FISHTRAC. Unsurprisingly, this resulted in mediocre performance on our DETRAC train set: just  62\% precision and 46\% recall (50.3 mAP).

\section{Experiment Setup}

\subsection{Evaluation Metrics}
\begin{figure}
\centering
\includegraphics[width=0.5\columnwidth]{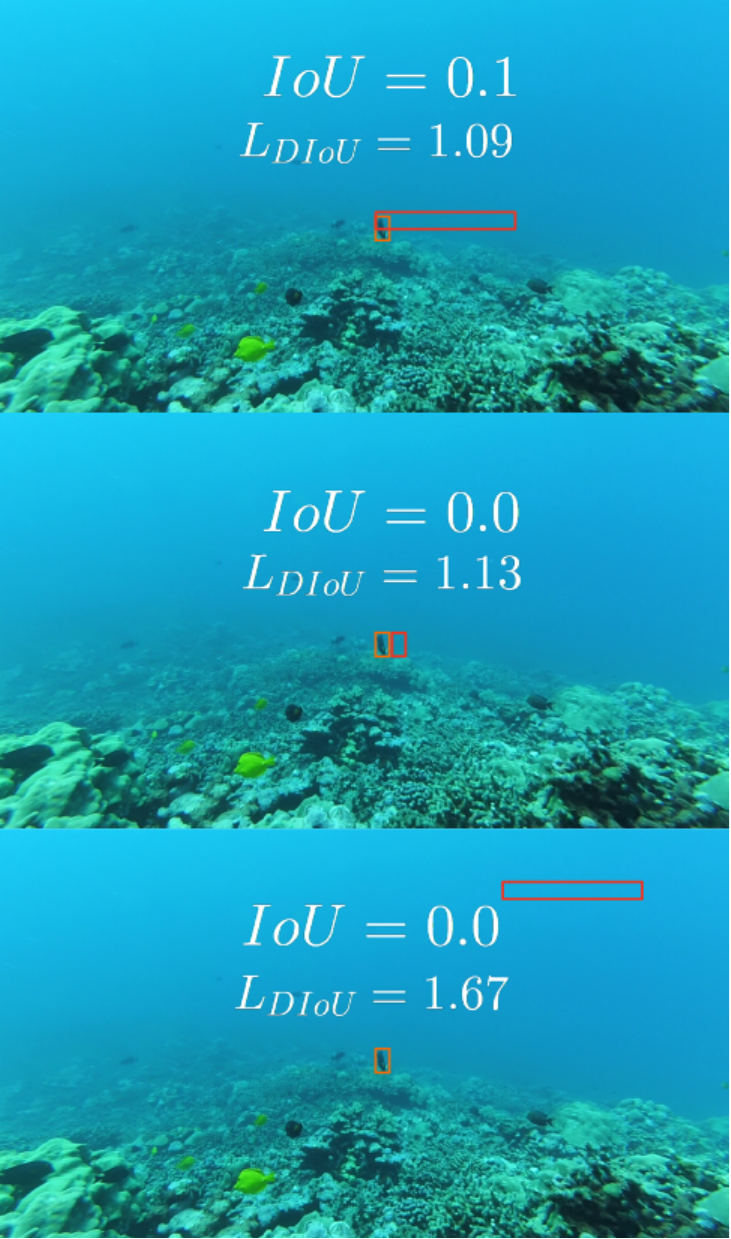}  
\caption{The scores of DIoU and IoU in a real fish tracking scenario. The orange box is the ground truth and the red is the predicted (tracked) box.  The middle image seems best from an end-use perspective, but using IoU there is no way to differentiate it from the bottom image, which is clearly much worse. }
\label{fig:diou}
\end{figure}

As a primary metric we use the recent HOTA metric~\cite{luiten2020hota}, which has gained popularity due to its strong performance in user studies~\cite{luiten2020hota}.  But we also report more classic  CLEAR MOT metrics like MOTA~\cite{bernardin2008evaluating}, as secondary metrics.

However, one limitation of these MOT evaluation metrics is that they are both based on the IoU (intersection over union) between each box in the predicted track and each box in the ground truth track.   MOTA uses a fixed threshold on IoU to determine whether a ground truth and a predicted box are close enough to match, whereas recent improvements like HOTA take the average score over many possible thresholds.  However, regardless of the exact threshold, if the predicted and the ground truth track do not overlap, the IoU value is zero. So a predicted box which does not overlap any ground truth box will count as a false positive.  Indeed, in such situations the tracker would have received a better score if it had simply not tracked the object at all.  The rationale behind this approach is that a tracker which completely loses track of the object should detect that it has failed and not issue a prediction so as not to confuse the downstream pipeline.

Although at first this seems reasonable, in our setting we found that this produced highly counter-intuitive results.  Consider Figure~\ref{fig:diou} for instance.  The middle image is clearly better than the top image at giving the user a sense of where the fish is: it has roughly the right size box in roughly the right location, whereas with the top image the location and size of the target are both completely inaccurate.  Yet with a low enough IoU threshold, the top image will count as a matched detection due to the overlap, while the middle image never will no matter the IoU threshold. Additionally, with the IoU metric, there is no way to differentiate the middle image from the bottom image, even though the middle prediction is clearly much more useful to an end-user than the bottom prediction.  With low-accuracy detections, situations similar to the middle image will happen a lot: especially when targets become small (such as fish swimming away from the camera) the tracker may have to rely on a motion model rather than visual information  to determine where the object is.  In these situations, as long as the tracker produces a track that is ``close'' to the original it will still be helpful for downstream applications even if there is no overlap, especially for small targets.  

Therefore, we instead use Distance-IoU (DIoU)~\cite{zheng2020distance}, a recent metric that combines IoU with the normalized distance between the boxes to give more ``partial credit'' to non-overlapping detections.
Specifically, DIoU computes
\begin{equation}
DIoU(b_1, b_2) = 1 - IoU(b_1,b_2) + \frac{d_{euclid}^2(C(b_1), C(b_2)}{g^2(b_1,b_2))},
\end{equation}
 where g is the diagonal length of the smallest box enclosing the two boxes and $C$ is the center point operator.  Intuitively, this combines IoU with the normalized distance between the boxes.   DIoU ranges from between 0 to 2 (note that, unlike IoU, lower is better), and we wish to have a threshold greater than 1 but less than 2 to admit boxes that may not overlap.  
If two equally-sized boxes barely touch at a corner, they will have DIoU 1.25, so this is the value we use to initialize the track.  DETRAC's CLEAR MOT implementation originally allowed 20\% variability in the threshold to allow for more leeway while tracking the object; we followed this approach, allowing the DIoU to rise up to 1.5 while tracking an object.  To ensure our HOTA and MOTA metrics were considering a similar range of DIoU values, we modified HOTA to integrate over DIoU values between 1.25 and 1.5.  The resulting scores on the training set of FISHTRAC more closely matched our intuition than the IoU-based scores, for instance DIoU with these thresholds would successfully count the middle scenario in Figure~\ref{fig:diou} as a match while excluding the much worse bottom scenario.  


\subsection{Evaluation Protocol}

Trackers fed low-accuracy detections might take an extremely long time, or might fail to produce any results. To handle the time issue, our code kills the tracker after 30 minutes have passed on a single video - this is recorded as a \textbf{timeout}.  In contrast, a \textbf{failure} occurs when a tracker fails to produce any results at all for an entire video, usually due to an assumption in the original code not being met - e.g. assuming that there are detections on every frame. 

Additionally, each tracker other than RCT requires setting $h$, the threshold on detection confidence.  We set this separately for each tracker. A robust tracker should never timeout or fail, so we first select the threshold(s) that minimize the sum of timeouts and failures.  In the case of ties, we select based on average HOTA over the DETRAC and FISHTRAC train sets.  We then use this threshold on the test videos.

After this process, all trackers had zero timeouts/failures on the training data, except for the three slowest methods (D3S, GMMCP, and IHTLS), which often timed out.

\subsection{RCT and Baseline Implementations}

\noindent\textbf{RCT Implementation Details:} Like most other MOT algorithms, RCT has a number of parameters. In our case, other than $h_I=0.5$ which was set purely based on intuition, we set the other 10 parameters to maximize MOTA and qualitative performance on the 6 FISHTRAC/DETRAC training videos. This resulted in the following settings:  $\beta=50\%$, $\delta = 4$, $\delta_m = 2$, $h_u = 0.3$,  $D_{max} = 20$, $h_q = 0.8$,  $h_f = 0.2$, $\omega=1\%$, $\alpha=1.1$, and $\delta_n = 5$. The majority (7 of 10) of these parameters control the trimming and joining heuristics (see ablation study for the impact of these components). A list of the RCT parameters used in our experiments are provided in Table \ref{tab:rct-params}. In terms of Kalman filter parameters, we set the transition and observation covariance matrices to standard 1-diagonal form (with 0 elements for the velocity observations since they are unobserved), although we did a small amount of tuning on the diagonal velocity transition elements (which were set to 0.2), and the diagonal position observation elements (which were set to 0.5). 

\begin{table}
	\begin{center}
	\caption{RCT parameters and meaning.}
    	\label{tab:rct-params}
	\resizebox{\columnwidth}{!}{
    	\begin{tabular}{|l|l|} 	
		\hline
		Parameter& Meaning\\
		\hline
		$h_i$&  Detection confidence threshold.\\
		\hline
		$\beta$& Percent box is enlarged to check if it is sufficiently far from image edge.\\
		\hline
		$\delta$& Number of previous frames used to calculate approximate position and velocity using Kalman filter.\\
		\hline
		$\delta_m$& Number of boxes needed to justify switching from Kalman filter to median flow.\\
		\hline
		$h_u$& IoU threshold to determine if two detections are potentially on the same object \\
		\hline
		$D_{max}$& Maximum number of frames with missing detections to permit joining two tracklets.\\
		\hline
		$h_q$& Detection confidence threshold used to filter ``high-quality" detections. \\
		\hline
		$h_f$& IoU threshold used to filter redundant tracks.  \\
		\hline
		$\omega$& Percentage offscreen an object must be in order to trim its track.  \\
		\hline
		$\alpha$& Acceleration factor when objects are moving offscreen. \\
		\hline
		$\delta_n$& Number of frames of missing detections needed before deciding to trim them from track. \\
		\hline
	\end{tabular}}
	\end{center}
\end{table}

\noindent\textbf{Classic and Specialized Baselines:} In total, we compare RCT to 15 trackers. We compare to four classic trackers from the original DETRAC set (\textbf{GOG}~\cite{pirsiavash2011globally}, \textbf{CMOT}~\cite{bae2014robust}, \textbf{RMOT}~\cite{yoon2015bayesian}, and \textbf{IHTLS}~\cite{dicle2013way}). To this we add \textbf{GMMCP}~\cite{dehghan2015gmmcp}, a tracker used in recent video-based person re-identification systems~\cite{liu2019spatial,jiang2021ssn}.  We compare two related improvements of the IOU tracker~\cite{bochinski2017high}, \textbf{KIOU} (which uses a Kalman Filter) and \textbf{VIOU}~\cite{bochinski2018extending}  (which uses MedianFlow). We compare \textbf{JPDA\_m}~\cite{rezatofighi2015joint}, an optimization of the classic JPDA approach~\cite{fortmann1983sonar} that, like RCT, incorporates motion model probability. We also compare to Visual Fish Tracking (\textbf{VFT})~\cite{jager2017visual}, which is specially designed to track fish in real-world video. 

\noindent\textbf{Deep Baselines:} We compare to \textbf{DAN}~\cite{sun2019deep}, which has exceptional performance on UA-DETRAC. We fine-tuned DAN on our train set (pretraining on the provided pedestrian model) to maximize performance.  We also compare \textbf{AOA}~\cite{du2020tao}, which won the recent 2020 ECCV TAO challenge and uses an improved version of the popular DeepSORT~\cite{wojke2018deep} algorithm.

\noindent\textbf{SOT Baselines:}  Comparing to SOT approaches is unusual in the MOT literature; however, SOT approaches rely less on detection quality and thus may be a viable approach in this setting.  We adapt these approaches to the MOT setting in a way that mirrors RCT: we initialize the tracker on the highest confidence detection that does not overlap previous tracks, and run the tracker forward and backward from that frame; continuing to add tracks while there are still uncovered detections. As SOT trackers, we try \textbf{MedianFlow}~\cite{kalal2010forward} and \textbf{KCF}~\cite{henriques2014high}, which have shown good performance in MOT pipelines~\cite{bochinski2018extending}. We also compare \textbf{D3S}~\cite{lukezic2020d3s}, a deep segmentation approach which is one of the top performers on the recent real-time VOT-RT 2020 challenge~\cite{kristan2020vot}. However, even ``real-time'' SOT trackers may be slow when applied to the more complex MOT task. Therefore, we compare to \textbf{GOTURN}~\cite{held2016learning}, a deep tracker which ranked \#1 in terms of speed and \#6 of 39 in accuracy on the large-scale GOT-10k benchmark~\cite{huang2019got}.


\section{Results}
\subsection{RCT Performance Analysis}
\begin{figure}
\centering
  \includegraphics[width=0.6\columnwidth]{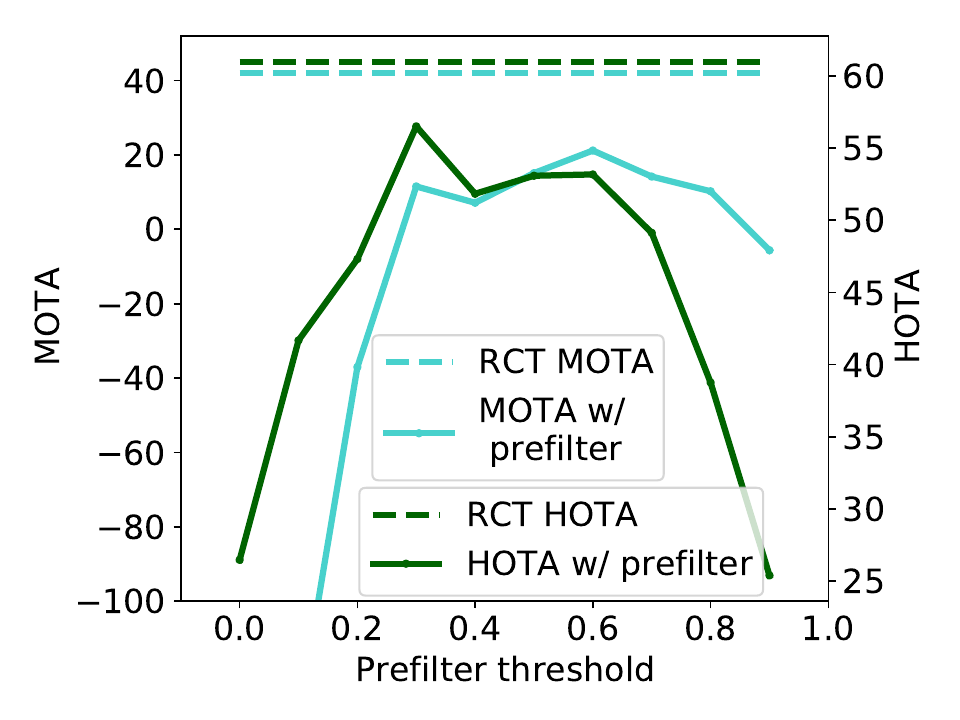} 
\caption{RCT  on FISHTRAC train with/without a prefilter.}
\label{fig:prefilter}
\end{figure}

One of the key aspects of RCT is its use of the exact detection confidence, instead of the standard method of ``prefiltering'' the detections by a fixed confidence threshold, and then discarding the confidence. Figure~\ref{fig:prefilter} shows a comparison of RCT to a variant with an initial prefilter.  No matter how we set the threshold, we cannot reach the original performance, showing the benefit of utilizing the exact detection confidence when tracking.
\\
\indent RCT is capable of efficiently searching through an unfiltered set of detections to produce an effective track.  Table~\ref{tab:unfiltered} shows that this is not the case for other trackers. Several methods could not cope with the large number of unfiltered detections, being unable to return a result even after three days.\footnote{The methods were either run on a state-of-the-art high-performance computing cluster, or on a modern GPU-capable server, depending on their hardware and software requirements.} 
The methods that completed showed poor performance. In contrast, RCT was able to quickly produce accurate tracks in this setting.

\begin{table}
  \centering
    \caption{Performance when trackers are fed unfiltered detections for one DETRAC video (MVI\_40752, 2025 frames). }
    \label{tab:unfiltered}
		\resizebox{0.8\columnwidth}{!}{
    \begin{tabular}{ l |l|l|l|l} %
      \textbf{Tracker} & \textbf{Time}  & \textbf{HOTA} & \textbf{ID switches} & \textbf{MOTA}  \\\specialrule{2.5pt}{1pt}{1pt}
		RCT & 13 min  & 51.44 & 8  & 36.78\\ \hline
		MEDFLOW & 36 min  & 45.47 & 88 &  -8.21 \\ \hline
		KCF & 57 min   & 44.89 & 307 & 2.13 \\ \hline
		DAN & 61 min   & 13.38 & 843 & -757.63 \\ \hline
		VIOU & 155 min  & 12.49 & 219 & -984.70 \\ \hline
		KIOU & 334 min  & 6.06 & 330  & -4126.46\\ \hline
		GOG & 452 min  & 15.40 & 367 & -828.10 \\ \hline
		GOTURN & 524 min  & 2.84 & 189 & -5499.44 \\ \hline
		AOA & 677 min  & 6.74 & 572 & -2219.49 \\ \hline
		D3S & 1010 min  & 10.09 & 125 & -951.38 \\ \hline
		JPDA\_m & 2483 min  & 9.50 & 835 & -1688.34 \\ \hline
		VFT & $>$ 3 days &  -- & -- & -- \\ \hline
		CMOT & $>$ 3 days &  -- & -- & -- \\ \hline
		RMOT & $>$ 3 days &  -- & -- & -- \\ \hline
		GMMCP & $>$ 3 days &  -- & -- & -- \\ \hline
		IHTLS & $>$ 3 days &  -- & -- & -- \\ \hline

    \end{tabular}}
  
\end{table}

\label{sec:ablate}
Given that our method contains several non-essential components, we ran an ablation study to determine the impact of each factor.  
The results are shown in Table~\ref{tab:ablation}.  We see that removing any of the various features of RCT does result in a decrease in the training data HOTA and MOTA, and typically an increase in ID switches.  
It was surprising that the precise method of track trimming had such a large impact - this may point to a deficiency in the HOTA/MOTA metrics as the differences in track trimming method often cause little to no visually noticeable change in the results.

\setcounter{table}{5}
\begin{table*}
\centering
    \caption{Test set results for FISHTRAC (shorthand: Fish) and DETRAC (shorthand: Car). Timeouts and Failures are summed across the datasets, while the Avg HOTA and Avg FPS are averaged. The table is sorted by the sum of timeouts and failures, and second by the average HOTA. Bolded values indicate the best scores of trackers that produced results on all sequences.  }
    \label{tab:test}
		\resizebox{\textwidth}{!}{
     \begin{tabular}{ l |p{1cm}|p{0.8cm}|p{1cm}|p{1cm}|p{1cm} |p{1.2cm} |p{1cm} |p{1cm} |p{1cm} |p{1cm}  |p{1cm} |p{1cm} |p{1cm} |p{1cm} |p{1cm}|p{1cm} } %
      \textbf{Tracker} & \textbf{Time-outs} & \textbf{Fail-ures} & \textbf{Avg HOTA} & \textbf{Total ID Sw} & \textbf{Fish HOTA} & \textbf{Fish MOTA} & \textbf{Fish ID Sw} & \textbf{Fish Prcn} & \textbf{Fish Recall}  & \textbf{Car HOTA} & \textbf{Car MOTA}  & \textbf{Car ID Sw} & \textbf{Car Prcn} & \textbf{Car Recall} & \textbf{Avg FPS} \\\specialrule{2.5pt}{1pt}{1pt}
		RCT & \textbf{0} &  \textbf{0} &  \textbf{44.58} &  \textbf{553} &  \textbf{49.67} &  \textbf{45.97} &  \textbf{47} &  83.65 &  57.48 &  39.49 &  29.60 &  \textbf{506} &  \textbf{94.15} &  31.64 &  4.08 \\ \hline 
		KCF &  \textbf{0} &  \textbf{0} &  43.16 &  3563 &  30.45 &  27.75 &  884 &  69.62 &  58.35 &  \textbf{55.87} &  \textbf{47.86} &  2679 &  81.62 &  62.33 &  20.68\\ \hline 
		MEDFLOW &  \textbf{0} &  \textbf{0} &  42.95 &  800 &  32.03 &  -58.51 &  108 &  37.01 &  \textbf{82.45} &  53.87 &  33.67 &  692 &  68.12 &  63.49 &  2.51\\ \hline
		DAN & \textbf{0} &  \textbf{0} &  42.02 &  17253 &  44.24 &  42.05 &  361 &  \textbf{90.73} &  49.17 &  39.80 &  35.00 &  16892 &  76.38 &  54.53 &  5.08\\ \hline
		GOG &  \textbf{0} &  \textbf{0} &  39.41 &  15873 &  37.85 &  45.08 &  414 &  87.48 &  55.42 &  40.98 &  39.42 &  15459 &  74.27 &  64.06 &  \textbf{94.17}\\ \hline
		AOA & \textbf{0} &  \textbf{0} & 35.21 & 20848 & 39.28 & 13.79 & 593 & 57.57 & 65.53 & 31.14 & 3.60 & 20255 & 52.25 & \textbf{79.00} & 10.40 \\ \hline
		\rowcolor{Gray}
KIOU &  0 &  1 &  46.64 &  5109 &  49.47 &  46.72 &  119 &  88.22 &  54.72 &  43.81 &  31.64 &  4990 &  65.98 &  66.95 &  159.45\\ \hline
\rowcolor{Gray}
VIOU &  0 &  1 &  45.86 &  2768 &  48.91 &  46.44 &  51 &  93.69 &  50.12 &  42.81 &  35.39 &  2717 &  71.98 &  58.64 &  5.41\\ \hline
 \rowcolor{Gray}
 JPDA\_m & 0 & 1 & 37.79 & 1357 & 34.11 & 35.75 & 77 & 94.69 & 38.35 & 41.47 & 32.99 & 1280 & 89.13 & 37.81 & 17.30 \\ \hline
\rowcolor{Gray}
	GOTURN &  0 &  2 &  20.47 &  1102 &  19.53 &  -281.09 &  114 &  16.24 &  67.47 &  21.41 &  -88.88 &  988 &  23.00 &  37.80 &  8.61 \\ \hline
	\rowcolor{Gray}
	RMOT &  5 &  0 &  38.28 &  1077 &  39.74 &  40.21 &  133 &  90.91 &  45.54 &  36.82 &  25.77 &  944 &  88.84 &  29.65 &  5.05\\ \hline
		\rowcolor{Gray}
		CMOT &  5 &  1 &  46.61 &  5731 &  54.40 &  50.30 &  110 &  84.47 &  62.41 &  38.82 &  12.71 &  5621 &  55.75 &  65.92 &  2.38\\ \hline
		\rowcolor{Gray}
		VFT &  8 &  0 &  23.49 &  5257 &  30.73 &  33.93 &  449 &  93.27 &  39.39 &  16.25 &  16.45 &  4808 &  90.55 &  19.22 &  12.23\\ \hline
		\rowcolor{Gray}
		D3S & 26 & 1 & 34.11 & 82 & 54.72 & 23.20 & 33 & 60.94 & 65.12 & 13.50 & 2.15 & 49 & 64.20 & 4.89 & 0.62 \\ \hline
		\rowcolor{Gray}
		GMMCP &  30 &  11 &  17.25 &  138 &  29.76 &  31.10 &  114 &  89.78 &  35.84 &  4.75 &  0.40 &  24 &  87.03 &  0.48 &  0.17\\ \hline
		\rowcolor{Gray}
		IHTLS &  45 &  1 &  6.80 &  242 &  13.60 &  -2.54 &  242 &  48.40 &  17.16 &  0.00 &  0.00 &  0 &  -- &  0.00 &  0.08\\ \hline


    \end{tabular}}
\end{table*}

\setcounter{table}{4}
\begin{table}
  \centering
    \caption{Ablation study. HOTA and MOTA are averaged over the two train datasets; ID switches are summed.}
    \label{tab:ablation}
		\resizebox{\columnwidth}{!}{
    \begin{tabular}{ p{4cm} |l|l|p{2cm}} %
      \textbf{Variation} & \textbf{Avg HOTA} & \textbf{Total ID Switches} & \textbf{Avg MOTA} \\\specialrule{2.5pt}{1pt}{1pt}
		Unmodified &  60.61 & 32 & 45.86 \\ \hline
		No MedianFlow  & 56.02 & 40 & 41.33 \\ \hline
		No track joining  & 58.32 & 46 & 45.27 \\ \hline
		Not filtering long, large, low confidence tracks  & 56.79 & 64 & 29.03 \\ \hline
		Not trimming when box is offscreen  & 25.61 & 38 & -608.37 \\ \hline
		Trimming as soon as box touches offscreen  & 57.06 & 23 & 36.65 \\ \hline
		Not trimming when box is fully onscreen  & 58.37 & 34 & 32.91 \\ \hline
    \end{tabular}}
\end{table}

\subsection{Test Results}

We ran all 16 trackers across all 11 FISHTRAC test videos and the 40 UA-DETRAC test videos. We followed good practice regarding test data, in particular, we did not in any way evaluate RCT on the test videos during its development.  Our objective is the same as it was when selecting thresholds: we wish to minimize timeouts and failures for reliability, and then to maximize the average HOTA score.

Test results are shown in Table~\ref{tab:test}.  Our main result is that, of the trackers which successfully produced results for every sequence (i.e. no timeouts or failures), our RCT algorithm has the best average HOTA across the FISHTRAC and the DETRAC dataset.  This demonstrates the advantages of RCT in terms of robust performance (which is notable given that RCT was developed based on  examining just 6 videos).
Many other trackers were not nearly as robust - for instance, while CMOT has an impressive HOTA score on the FISHTRAC dataset, it cannot cope with the longer DETRAC sequences, resulting in 5 timeouts and 1 failure.  In contrast, our adaptation of the KCF single-object tracker does extremely well on DETRAC, but significantly worse on FISHTRAC - likely because fish are significantly more difficult to track based on visual information due to appearance changes etc.  The fact that KCF and MEDFLOW perform so well on DETRAC highlight the importance of comparing to SOT algorithms even when attempting to solve a MOT problem. 
Although KCF is thought to be quite a low baseline on SOT problems, our experiments indicate that stronger SOT trackers like D3S are too computationally expensive to run on our MOT problems - in fact, D3S timed out on over half the test videos.  GOTURN has sufficient speed, but performs poorly, in part due to not adequately handling MOT-specific issues such as track termination.   

One of the most notable features of our RCT algorithm is how it achieves just 553 identity switches across all 51 test videos - the only algorithms with fewer are D3S, GMMCP, and IHTLS, algorithms that simply did not produce any tracks for the majority of videos.  The other MOT trackers have an order of magnitude  more identity switches, even algorithms such as DAN, VIOU, and KIOU which achieve good HOTA. This is due to RCT's ability to fuse low-confidence detections, a motion model, and a single object tracker to rapidly produce high-quality continuous tracks even when high-confidence detections are sparse. Minimizing ID switches is very important for practical applications - for instance, we intend to use RCT to help divers keep track of individual fish while underwater.  Numerous ID switches are likely to confuse the diver and cause them to follow the incorrect fish.  In these types of applications, we would much rather miss some objects, but ensure the tracks we do provide are high-quality, with little to no identity switches, even in the face of unreliable detections.   We expect RCT to excel in these situations.  

\section{Conclusion}
We have studied the problem of multi-object tracking-by-detection with unreliable detections. To illustrate this, we presented a new MOT dataset, FISHTRAC, with high-resolution videos of underwater fish behavior.  We also present RCT, which takes a different approach than other MOT algorithms, using the detection confidence in three different ways to produce high-quality tracks given completely unfiltered set of input detections. We find that RCT outperforms baselines (including the 2020 TAO challenge winner and a top performer on the VOT-RT 2020 challenge), tracking objects accurately with very few ID switches and no timeouts or failures. The public release of our FISHTRAC dataset and codebase are forthcoming.

A next step is adapting RCT to work in an online and real-time fashion in a way that can be deployed in the field. One practical benefit of RCT is that it does not use a GPU, which in edge settings may be fully utilized by the detection network.  Additionally,  we found many of high-performing MOT methods work poorly with a low-quality detector; so it would be interesting to explore an adaptive approach which analyzes detection quality and adapts the tracker behavior accordingly. Pursuing these directions will help MOT algorithms be more easily deployed to solve a diverse set of real-world problems.

\section{Acknowledgements}
We gratefully acknowledge the assistance of Timothy Kudryn, Ilya Kravchik, Dr. Timothy Grabowski, Christopher Hanley, and Sebastian J. Carter on this research project.  This work was supported by NSF CAREER Award \#HCC-1942229 and NSF EPSCoR Award \#OIA-1557349. 
\bibliography{kpdpaper}

\end{document}